\documentclass[10pt,journal,compsoc]{IEEEtran}

\usepackage{makecell}

\ifCLASSOPTIONcompsoc
  \usepackage[nocompress]{cite}
\else
  \usepackage{cite}
\fi

\ifCLASSINFOpdf
   \usepackage[pdftex]{graphicx}
   \graphicspath{{../../images/}}
   \DeclareGraphicsExtensions{.pdf,.jpeg,.png}
\else
\fi

\usepackage{amsmath,amssymb,amsfonts}

\usepackage{caption}

\newcommand\Tstrut{\rule{0pt}{2.25ex}}
\newcommand\TTstrut{\rule{0pt}{3.3ex}}

\begin{document}
\title{Distillation Strategies for \\ Proximal Policy Optimization}

\author{
	\IEEEauthorblockN{Sam Green\IEEEauthorrefmark{3}\IEEEauthorrefmark{1}, Craig M. Vineyard\IEEEauthorrefmark{3}, \c{C}etin Kaya Ko\c{c}\IEEEauthorrefmark{1}}\\
	\IEEEauthorblockA{\IEEEauthorrefmark{3}Sandia National Laboratories, Albuquerque, New Mexico, USA
		\\\{sgreen, cmviney\}@sandia.gov}\\
	\IEEEauthorblockA{\IEEEauthorrefmark{1}University of California, Santa Barbara, California, USA
		\\cetinkoc@ucsb.edu}
}

\IEEEtitleabstractindextext{%
\begin{abstract}
Vision-based deep reinforcement learning (RL) typically obtains performance benefit by using high capacity and relatively large convolutional neural networks (CNN). However, a large network leads to higher inference costs (power, latency, silicon area, MAC count). Many inference optimizations have been developed for CNNs. Some optimization techniques offer theoretical efficiency, such as sparsity, but designing actual hardware to support them is difficult. On the other hand, \textit{distillation} is a simple general-purpose optimization technique which is broadly applicable for transferring knowledge from a trained, high capacity teacher network to an untrained, low capacity student network. \textit{DQN distillation} extended the original distillation idea to transfer information stored in a high performance, high capacity teacher Q-function trained via the Deep Q-Learning (DQN) algorithm. Our work adapts the DQN distillation work to the actor-critic Proximal Policy Optimization algorithm. PPO is simple to implement and has much higher performance than the seminal DQN algorithm. We show that a distilled PPO student can attain far higher performance compared to a DQN teacher. We also show that a low capacity distilled student is generally able to outperform a low capacity agent that directly trains in the environment. Finally, we show that distillation, followed by ``fine-tuning'' in the environment, enables the distilled PPO student to achieve parity with teacher performance. In general, the lessons learned in this work should transfer to other modern actor-critic RL algorithms.
\end{abstract}

}

\maketitle

\IEEEdisplaynontitleabstractindextext

\IEEEpeerreviewmaketitle

\IEEEraisesectionheading{\section{Introduction}\label{sec:introduction}}

\IEEEPARstart{I}{n} 2013, DeepMind famously demonstrated above-human levels of performance on many Atari video games using the end-to-end deep reinforcement learning (RL) algorithm Deep Q-Learning (DQN) \cite{mnih_human-level_2015}. Since then, many improved RL algorithms have been developed. RL is currently being applied to diverse tasks, such as robotic manipulation, games, finance, medicine, and marketing. In the coming years, as RL applications continue to expand and task performance continues to increase, we foresee RL run-time efficiency becoming a critical issue. Demand for run-time efficiency will require special algorithmic considerations and ultimately specialized hardware. A parallel to this claim is the success and breadth currently enjoyed by the field of deep learning which has now entered a phase of hardware accelerator mass production by many industrial and academic groups.

\begin{figure}[!t]
\centering
\includegraphics[width=.8\linewidth]{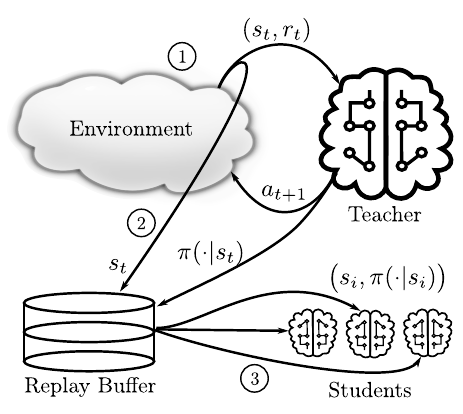}
\caption{RL distillation has three phases: 1) teacher training, using standard RL algorithms, 2) using the trained teacher to interact with the environment and saving state observations and teacher's action probabilities into a \textit{replay buffer}, and 3) the transfer of information stored in the replay buffer into student(s).}
\label{fig:teacher-student}
\end{figure}

\textit{Distillation} is a method to transfer information learned by a high capacity, high parameter-count teacher neural network into a relatively low capacity, low parameter-count student neural network \cite{hinton2015distilling}. When using distillation, all teacher output probabilities are used as a training signal for the student, versus the single label that is normally used during training. \textbf{Distillation leverages the fact that trained teacher class probabilities contain more information than a single label.} For example, if an apple is presented to a classifier successfully trained to recognize images of food, then the classifier's class probabilities for apple, pear, peach, and orange most likely have some significant value compared to non-round foods. Furthermore, the low probabilities for other non-round classes provide information about what the input is \textit{unlikely} to be. 

The results presented in this paper extend the work of \cite{rusu2015policy} which used distillation to train a student neural network to match the deep Q-network (DQN) of a teacher trained through the Deep Q-Learning algorithm. We refer to that technique as \textit{DQN distillation}. A noteworthy feature of DQN distillation, and any variety of RL distillation in general, is that only the teacher is required to experience the environment. Once trained, the teacher may pass its knowledge to students, without the students being required to experience the environment as well, Fig. \ref{fig:teacher-student}. Excellent results were obtained in \cite{rusu2015policy}, with the student DQNs often matching or exceeding the performance of the teacher DQNs on all tasks. 

Actor-critic algorithms constitute a popular family of high performance deep RL algorithms. In the context of deep RL, actor-critic algorithms are typically composed of two networks: an actor network, which also serves as the agent's policy, and a critic network, which serves as a value function during policy improvement. DQN only uses a value function, which is queried during run-time. In this work, we reexamine DQN distillation in the context of the Proximal Policy Optimization algorithm, which was developed more recently than DQN and subsequently has many improvements \cite{schulman_proximal_2017}. PPO was selected as our actor-critic algorithm because it is simple to implement and is widely used for a broad range of RL applications \cite{8461113, Rajeswaran_2018, zoph2017learning, 8462002}.

As RL becomes appropriate for real-world applications, various ``costs'' to execute neural network forward-propagation becomes critical. Action latency, power consumption, silicon area requirements, and other design factors must be reconciled with the fact that relatively large neural networks typically provide state of the art results. RL distillation techniques will be broadly useful for neural architecture design. In particular, RL distillation will allow a machine learning engineer to 1) design the best policy, given their hardware constraints, or 2) identify minimum hardware requirements, given a satisficing agent performance metric. RL distillation methods provide the following benefits:
\begin{itemize}
\item Rapid student model exploration is enabled by the use of an experience replay buffer. RL distillation keeps a large replay buffer, which is populated with high quality state observations, actions, and action probabilities recorded by the teacher after its training is complete.
\item Faster training times via high capacity teachers. It has been shown that high capacity agents decrease training time for both deep learning and deep RL \cite{rusu2015policy, 8489519}.
\item Expensive environments, e.g. accurate physics simulations or physical systems, may only need to be experienced once by the teacher. The teacher's replay buffer may then be repeatedly used for offline actor distillation at a later date.
\end{itemize}

\section{Background and Related Work}

\begin{figure*}
    \centering\
	\includegraphics[width=\textwidth]{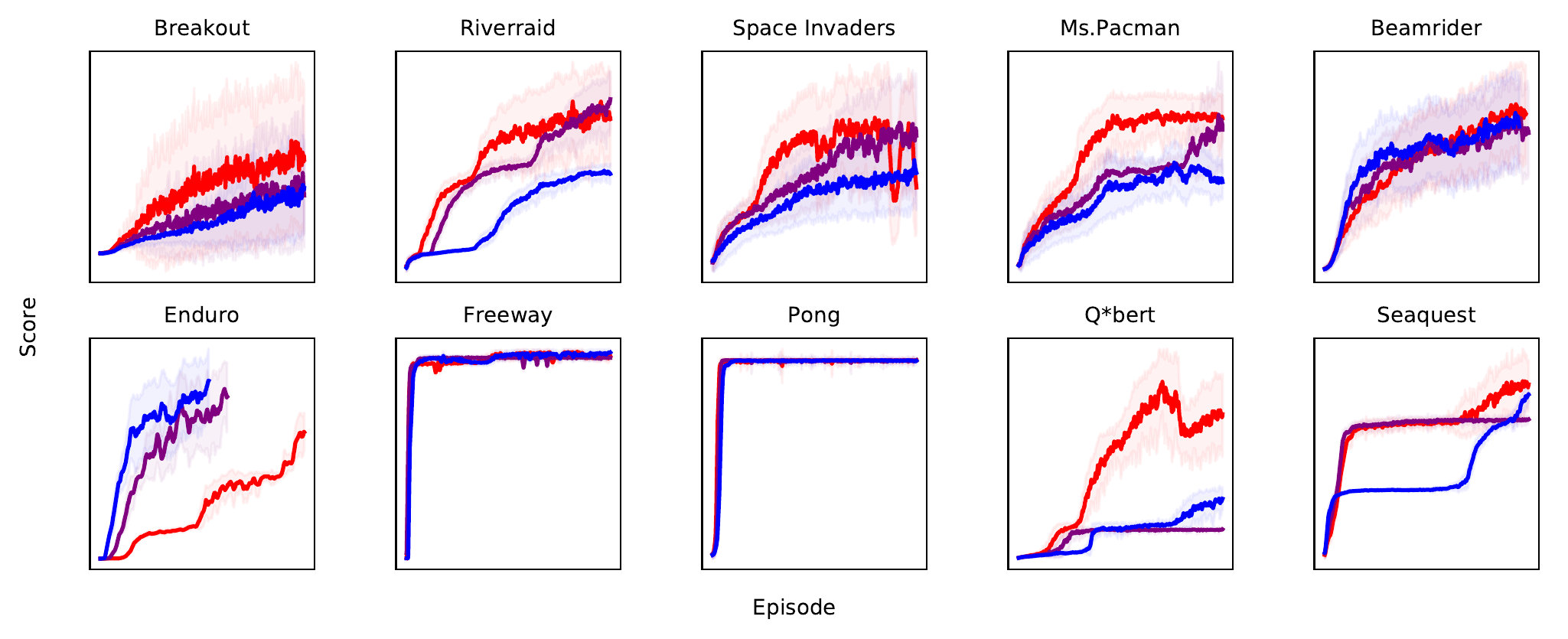}	
	\caption{Qualitative analysis of effect of policy capacity on learning rate. Red plot is high capacity, purple is medium capacity, and blue is low capacity. In general, high capacity policies achieved higher performance, faster than low capacity policies. Y-axis is game score, x-axis is number of games played. Total number of steps was fixed across all games to $75 \times 10^{6}$ time steps. As detailed in the Results section, distillation allows for higher performing low capacity policies, compared to what they may achieve through environmental interaction alone.}
    \label{fig:score-cap2}
\end{figure*}

Distillation was proposed in \cite{hinton2015distilling} as a method to transfer knowledge from a trained teacher classifier neural network into an untrained student network. There are various techniques to implement neural network distillation, and here we review the version most relevant to RL. Initially, assume a high capacity teacher classifier network has been trained to high performance, and a smaller network is to be trained with distillation. Additionally, assume access to the training inputs $X$ used for teacher training, but no access to class $C$ training labels $y \in C$. In this case, we may derive a loss function for the student network by providing training inputs $x \in X$ to the teacher network and using its class probability distribution $p_t(x) \in \mathbb{R}^{\lvert C \lvert}$ as a soft target for the student network's output probability distribution $p_\theta(x) \in \mathbb{R}^{\lvert C \lvert}$, where the student is parameterized by $\theta$. The student's loss is defined as the distance between distributions $p_t(x)$ and $p_\theta(x)$ and may be measured using a standard metric, such as Kullback-Leibler divergence: 
\begin{equation}
\mathcal{L}(p_\theta(x) \lvert p_t(x)) = \sum_{i=1}^{\lvert C \lvert} p_{t,i}(x) \,\text{log}\,\frac{p_{t,i}(x)}{p_{\theta,i}(x)},
\label{eq:dist3}
\end{equation}
where $p_{t,i}(x)$ and $p_{\theta,i}(x)$ represent the probability for class $i$, given input $x$. The gradient of $\mathcal{L}$ may then be taken with respect to the student's parameters, which may then be updated using gradient descent.

As introduced by \cite{rusu2015policy}, distillation without labels maps to the RL setting. In the context of value-based algorithms like DQN, the output of the teacher Q-network is a vector of state-action values $q_t(s) \in \mathbb{R}^{\lvert A \lvert}$, where $s$ is a state observation and $A$ is a discrete action space. A probability distribution $p_t(s)$ may be obtained from the teacher by taking the softmax of $q_t(s)$. The state observations are then also provided to the untrained student network parameterized by $\theta$, and its (originally random) state-action values may be interpreted as a probability vector by taking the softmax of its output, giving $p_{\theta}(s)$. The trained teacher Q-network is then distilled into a student network using the Kullback-Leibler divergence metric for the loss:
\begin{equation}
\mathcal{L}(p_\theta(s) \lvert p_t(s)) = \sum_{i=1}^{\lvert A \lvert} p_{t,i}(s) \,\text{log}\,\frac{p_{t,i}(s)}{p_{\theta,i}(s)},
\label{eq:dist1}
\end{equation}
where $p_{t,i}(s)$ and $p_{\theta,i}(s)$ represent the probability for action $i$, given state observation $s$. 

\begin{table}
	\centering
	\begin{tabular}{|r||c|c|c|c|} \hline
				    &	DQN        & PPO Teacher & PPO Medium & PPO Low \Tstrut\\ \hline
		Beamrider   & \bf{8672.4}   & 7500 		& 7018 	     & 6958 \Tstrut\\
		Breakout 	& \bf{303.9}	   & 277 	    & 166 	     & 187 \\
		Enduro 		& 475.6 		   & 722 	    & 827 	     & \bf{948} \\
		Freeway 	    & 25.8 	       & \bf{34} 	& 33 	     & \bf{34} \\
		Ms.Pacman 	& 763.5 		   & 3410 		& \bf{4544}  & 2085 \\
		Pong 		& 16.2 		   & \bf{21}	    & \bf{21}    & \bf{21} \\
		Q*bert 		& 4589.8 	   & \bf{28367} & 11646 	     & 18502 \\
		Riverraid 	& 4065.3 	   & 13916 	    & \bf{15601} & 9408 \\
		Seaquest 	& \bf{2793.3}  & 2471 		& 1908 	     & 2315 \\
		S. Invaders & 1449.7 	   & \bf{1653}	& 1624 	     & 1312 \\ \hline \hline
		\% of DQN   & 100\%         & 169\%      & 150\%      & 141\% \Tstrut\\ \hline 
	\end{tabular}
\caption{Comparisons of policies trained by DQN and PPO. DQN results are taken from \cite{rusu2015policy}. PPO Teacher, PPO Medium, and PPO Low refer to agents trained using PPO with high, medium and low capacity policies respectively. Policy details are given in the Implementation Details section. All PPO policies were trained for $75 \times 10^6$ environment time steps and evaluated for $1 \times 10^6$ time steps. ``\% of DQN'' is the geometric mean of the column divided by the geometric mean of the DQN column. DQN and PPO Teacher have the same architecture. Medium and low capacities have 25\% and 7\% of the parameters as DQN and PPO Teacher. Note that PPO Low has a geometric mean 41\% higher than DQN, and that higher capacity PPO policies tend to have higher performance than lower capacity PPO policies.}
\label{tab:dqn_baseline_comps}
\end{table}

Eq.\,\ref{eq:dist1} would lead to low agent performance if used as given for DQN distillation. Recall that Q-values represent the expected return from state $s$, given that action $a$ is taken, and the policy is followed thereafter. After training is complete, an agent makes its decisions by taking the action with the highest Q-value. By taking the softmax of the DQN, we are interpreting the Q-values as a probability distribution. This distribution may be relatively uniform, and because of the noise introduced during distillation, values in the student may not relatively match that of the teacher. Specifically, the Q-value for a suboptimal action in the teacher may become the highest Q-value in the student, and this would lead to degraded agent performance. The authors of \cite{rusu2015policy} minimized the chance of this error by dividing all teacher Q-values by a temperature parameter $\tau=.01$, prior to calculating the softmax. This has the effect of ``sharpening'' the teacher's probability distribution in $p_t(s)$, such that the highest probability is much greater than the next to highest.

After the teacher has been fully trained, a distillation training set is collected from the teacher into a replay buffer. The authors of \cite{rusu2015policy} showed excellent distilled student performance across a variety of classic Atari 2600 games. Most significantly, a low capacity student network, with 7\% of the parameters relative to their teacher network, performed at least as well as the teacher network.

PPO is an actor-critic algorithm which has stood out as being simple to implement and high-performance \cite{schulman_proximal_2017}. PPO is now established as a popular baseline with which to compare other RL algorithms and as a preferred algorithm for applying RL to new tasks and for applications outside RL algorithm research \cite{zhang2018vrgoggles, rajeswaran2017learning}. Because of the popularity and performance of PPO, it was selected as our actor-critic algorithm.

In general, PPO learns more efficiently than the seminal DQN algorithm. Table\,\ref{tab:dqn_baseline_comps} compares agents trained with DQN and PPO. Significantly, PPO agents with much smaller capacity (7\%) achieved 41\% higher than a high capacity DQN agent, when comparing geometric means. 

A motivating factor for policy distillation is that it may be used to increase the sample efficiency and optimize the performance of a low capacity policy. In \cite{rusu2015policy} it was speculated that a larger network accelerates learning. In \cite{8489519} it was observed that high capacity policies are generally able to learn a task better and faster than low capacity policies. In the context of this work, the results in Fig.\,\ref{fig:score-cap2} also show that high capacity policies have performance advantages. In this figure, the average scores for agents using three different policy architectures are tracked during training for $75 \times 10^6$ time steps. All PPO agents were trained using Proximal Policy Optimization, as described in the Implementation Details section.

Distillation has also proven to be useful for neuromorphic hardware design. For example, the benefits of better sample efficiency and higher student performance through distillation were combined in \cite{mckinstry2018low} for efficient RL policy development. In this work, a high capacity policy trained with Double DQN, and represented by a standard convolutional neural network (CNN), was distilled into a student policy represented by a low precision spiking neural network to be executed on IBM's TrueNorth architecture. As TrueNorth has special restrictions, e.g. binary activations and ternary weights, it does not use a standard SGD algorithm. Instead TrueNorth uses  the Energy-Efficient Deep Networks algorithm \cite{2016arXiv160308270E} to train a student to match a teacher's Q-values. Importantly, \cite{mckinstry2018low} demonstrates the viability of training a teacher policy once, using one type of algorithm, and distilling that policy into an arbitrary number of student policies, using the best training algorithm for each respective student.

\section{Formulation}
Actor distillation (AD) is an offline technique closest in formulation to DQN distillation, with the difference being that AD distills the teacher's true actor probabilities, i.e. the teacher's policy, $\pi_t$ into the student $\pi_\theta$, which are both functions of state observation $s$. Whereas DQN distillation transfers a proxy of the teacher's value function into the student.

AD proceeds as follows: a teacher policy $\pi_t$ is trained to maximum performance on the environment. After training, the trained teacher interacts with the environment during a collection phase which records the teacher's state observations and action probabilities to a replay buffer. An uninitialized student network $\pi_\theta$ is then trained to mimic the teacher with mini-batch SGD using the replay buffer and a loss similar to Eq.\,\ref{eq:dist1}:
\begin{equation}
\mathcal{L}(\pi_\theta(s) \lvert \pi_t(s)) = \sum_{i=1}^{\lvert A \lvert} \pi_t(a_i \lvert s) \,\text{log}\,\frac{\pi_t(a_i \lvert s)}{\pi_\theta(a_i \lvert s)},
\label{eq:dist2}
\end{equation}
where $s$ and $\pi_t(\cdot \lvert s)$ are stored in the replay buffer.

$\pi_t$ and $\pi_\theta$ are obtained by taking the softmax of a policy network logits vector. \cite{rusu2015policy} obtained better results by dividing the teacher logits by .01, prior to taking the softmax. This has the effect of sharpening the teacher's probabilities. This was necessary because Q-values, which are learned using an $\epsilon$-greedy explore-exploit strategy, have undefined behavior when converted to a distribution. AD does not require sharpening, because the teacher policy is stochastic anyway.

Optionally, after distillation is complete, the student may be fine-tuned by allowing it to interact directly with the environment and using a standard actor-critic algorithm.

\section{Implementation Details}

\begin{table}
	\centering
	\begin{tabular}{|r||c|c|c|c|c|} \hline
		Capacity & Layer & Channels & Shape & Stride \Tstrut\\ \hline
		High     & Conv 1     & 32       & 8     & 4      \Tstrut\\ 
		         & Conv 2     & 64       & 4     & 2      \\ 
		         & Conv 3     & 64       & 3     & 1      \\ 
		         & FC 1   & n/a      & 512   & n/a    \\ \hline

		Medium   & Conv 1     & 16       & 8     & 4      \Tstrut\\ 
		         & Conv 2     & 32       & 4     & 3      \\ 
		         & Conv 3     & 32       & 3     & 1      \\ 
		         & FC 1    & n/a      & 256   & n/a    \\ \hline

		Low      & Conv 1     & 8        & 8     & 4      \Tstrut\\ 
		         & Conv 2     & 16       & 4     & 2      \\ 
		         & Conv 3     & 16        & 3     & 1      \\ 
		         & FC 1   & n/a      & 128   & n/a    \\ \hline
	\end{tabular}
\caption{Architecture details of policy feature extraction layers. High, medium, and low capacity agents were trained to play Atari. All architectures had three convolutional layers with inputs of $84 \times 84 \times 4$, followed by two fully connected layers (FC 1), followed by separate ``heads'': a fully-connected policy layer ($\pi$) with 3--18 units, depending on the environment, and a critic ($V$) unit with 512 units. The high, medium, and low capacity architectures had ``bodies'' with 1683456, 422912 (25\%), and 106752 (6\%) parameters, respectively.}
\label{tab:arch}
\end{table}

\begin{figure}
    \centering\
	\includegraphics[width=1.0\linewidth]{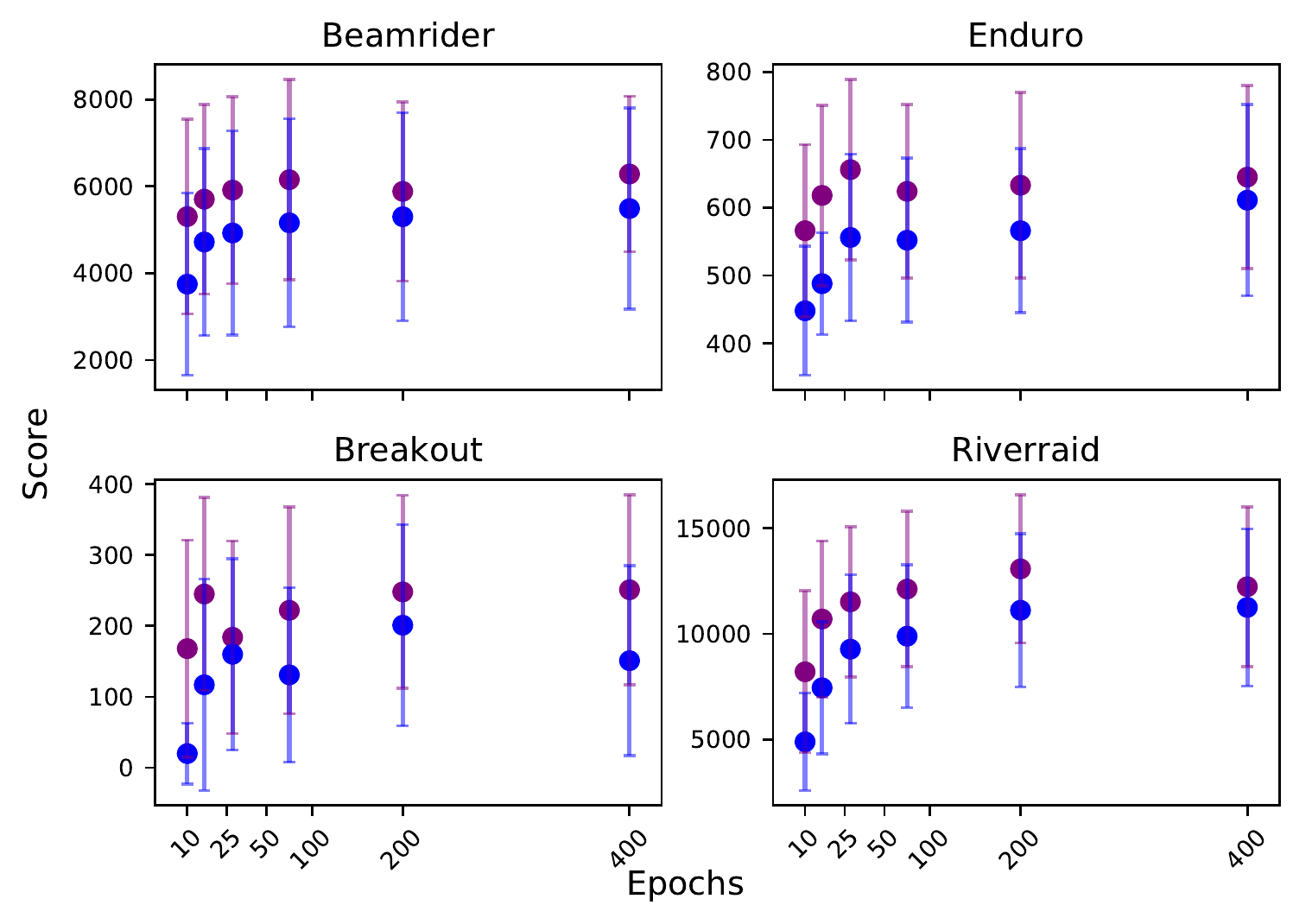}	
	\caption{Effects of capacity on distilled performance. Student policies were distilled from between 10 and 400 epochs and then evaluated for $1 \times 10^6$ time steps. Purple represents medium capacity policies and blue represents low capacity policies. Policies were reinitialized and distilled for each data point. Medium capacity policies have an advantage over low capacity policies when fewer epochs are used, but the advantage is often reduced as distillation progresses. As epochs increase, policies converge to maximum scores.}
    \label{fig:epochs}
\end{figure}

In this work, actor distillation was used to train students on 10 different Atari environments. We analyze the effect of AD on student performance, compared to agents with the same policy architecture as the student but trained directly on the environment with no distillation. We also analyze the effect of capacity on student performance, relative to agents with the same capacity but trained directly in the environment. Finally, we study the impact of allowing a distilled student to fine-tune on the environment after distillation is complete. Our environments are provided by the Arcade Learning Environment \cite{bellemare13arcade} and are interfaced with OpenAI Gym \cite{gym}. Additionally, we used PPO and distillation reference codes from \cite{pytorchrl} and \cite{2018arXiv180205668P}.

\begin{table*}
	\centering
	\begin{tabular}{|r||c|c|c|c|c|c|c|} \hline
	      & \makecell{Teacher}                    & Medium                               & Medium AD              			&    Medium AD tuned       & Low                                      & Low AD                     &        Low AD tuned \Tstrut\\ \hline
Beamrider & \makecell{15548\\ \bf{7500$\pm$2322}} & \makecell{12464\\7018$\pm$2183} & \makecell{(400) 11720\\6284$\pm$1790} & \makecell{10278\\5447$\pm$1867} & \makecell{12174\\6958$\pm$1997}          & \makecell{(400) 10974\\5489$\pm$2313} & \makecell{9744\\5548$\pm$1973} \TTstrut\\ \hline
Breakout  & \makecell{442\\277$\pm$115}     & \makecell{395\\166$\pm$123}     & \makecell{(200) 425\\248$\pm$136}     &  \makecell{434\\\bf{309$\pm$127}} & \makecell{372\\187$\pm$69}              & \makecell{(200) 414\\201$\pm$142} & \makecell{397\\159$\pm$103} \TTstrut\\ \hline
Enduro    & \makecell{983\\722$\pm$110}           & \makecell{1288\\827$\pm$223}    & \makecell{(50) 967\\656$\pm$133}      & \makecell{1062\\733$\pm$217} & \makecell{1386\\ 948$\pm$181}     & \makecell{(400) 791\\611$\pm$141} & \makecell{1376\\\bf{1013$\pm$162}} \TTstrut\\ \hline
Freeway   & \makecell{34\\34$\pm$1}               & \makecell{33\\33$\pm$0}         & \makecell{(10) 34\\34$\pm$1}          & \makecell{32\\32$\pm$0} & \makecell{34\\ \bf{34$\pm$0}}          & \makecell{(10) 34\\33$\pm$1} & \makecell{33\\33$\pm$0} \TTstrut\\ \hline
Ms.Pacman & \makecell{3940\\3410$\pm$333}         & \makecell{8140\\ 4544$\pm$875} & \makecell{(200) 4920\\3413$\pm$355} & \makecell{6500\\\bf{5041$\pm$1263}} & \makecell{2090\\2085$\pm$67}          & \makecell{(100) 4830\\3390$\pm$407} & \makecell{5450\\3483$\pm$302} \TTstrut\\ \hline
Pong      & \makecell{21\\21$\pm$2}               & \makecell{21\\21$\pm$1}         & \makecell{(10) 21\\20$\pm$3}           & \makecell{21\\19$\pm$7} & \makecell{21\\ \bf{21$\pm$0}} & \makecell{(10) 21\\19$\pm$7} & \makecell{21\\\bf{21$\pm$0}} \TTstrut\\ \hline
Q*bert    & \makecell{30075\\28367$\pm$3651}      & \makecell{11675\\11646$\pm$377} & \makecell{(10) 29975\\ \bf{28554$\pm$3301}} & \makecell{22100\\20572$\pm$2209} & \makecell{20775\\18502$\pm$2563} & \makecell{(10) 29650\\23019$\pm$9190} & \makecell{15375\\12152$\pm$1208} \TTstrut\\ \hline
Riverraid & \makecell{18980\\13916$\pm$3552}      & \makecell{20960\\ \bf{15601$\pm$3189}} & \makecell{(200) 18830\\13080$\pm$3517} & \makecell{18640\\13716$\pm$2874} &  \makecell{9910\\9408$\pm$217}   & \makecell{(400) 18630\\11256$\pm$3721} & \makecell{18430\\15593$\pm$2674} \TTstrut\\ \hline
Seaquest  & \makecell{4580\\2471$\pm$452}         & \makecell{1980\\1908$\pm$56}    & \makecell{(10) 4980\\2572$\pm$580} & \makecell{6060\\\bf{3708$\pm$1144}} & \makecell{2480\\2315$\pm$103}       & \makecell{(10) 4100\\2219$\pm$465} & \makecell{5940\\3550$\pm$1055} \TTstrut\\ \hline

Space Invaders & \makecell{2775\\ \bf{1653$\pm$451}} & \makecell{3025\\1624$\pm$596} & \makecell{(200) 2550\\1432$\pm$469} & \makecell{2550\\1569$\pm$375} & \makecell{2325\\1312$\pm$336}            & \makecell{(200) 2405\\1382$\pm$343} & \makecell{2205\\1288$\pm$263} \TTstrut\\ \hline \hline
\% of Teacher & 100\% & 88\%   &  94\%  & 100\% & 84\%  & 85\%  & 89\% \Tstrut\\ \hline 
\% of DQN     & 169\% & 150\%  &  160\% & 169\% & 141\% & 144\% & 151\% \Tstrut\\ \hline 
\end{tabular}
\caption{Game scores over $1 \times 10^6$ time steps using directly-trained (Teacher, Medium, Low columns), distilled (Medium AD, Low AD), and distilled-then-tuned agents (Medium AD tuned, Low AD tuned). Teachers use high capacity architecture trained with PPO for $75 \times 10^6$ time steps. Medium and Low represent medium and low capacity architectures. AD represents actor distillation as described in Implementation Details section. The top row in each cell is agent's high score over all games. The bottom row in each cell provides mean game score $\pm$ standard deviation. Parentheses in AD columns provide number of epochs used for student distillation. ``\% of Teacher'' columns are the geometric mean of column mean divided by geometric mean of teacher. ``\% of DQN'' provides geometric means of students versus those obtained through DQN distillation \cite{rusu2015policy}. Of note, medium capacity distilled student achieved 94\% of PPO-based teacher, and 160\% of DQN-based teacher results, which were, in turn, often higher than human player scores \cite{mnih_human-level_2015}. Medium capacity students distilled for 10 epochs and then allowed to train using PPO directly on the environment for $20 \times 10^6$ time steps (``Medium AD tuned'') achieved Teacher's level of performance. Similar analysis applies to low capacity students.}
\label{tab:avg_scores}
\end{table*}

Our PPO architectures use a single convolutional neural network body, followed by two separate ``heads'': one for the actor and one for the critic. Two student capacities were investigated: one with medium capacity and one with low capacity, both relative to the teacher's high capacity network. In order to make a fair comparison, student network architectures were chosen to match those used for the DQN Distillation results. Network architecture details are given in Table\,\ref{tab:arch}. 

For distillation and architecture baseline comparisons, the high (teacher), medium, and low capacity agents were trained for $75 \times 10^6$ time steps on each Atari environment. 16 agents ran in parallel with 2048 environment steps on each agent between each PPO update. Generalized Advantage Estimation was used to calculate returns with $\gamma=.99$ and $\tau=.95$. Within PPO, 10 epochs were used with batch sizes of 32 and clipping parameter set to .1. Adam was used with the stepsize set to $3 \times 10^{-4}$. Unlike \cite{rusu2015policy}, we do not divide teacher probabilities by a temperature and therefore directly use Eq.\,\ref{eq:dist2} for distillation. Tuning experiments on Beamrider, Enduro, Breakout, and Riverraid were used for final hyperparameter selection. 

\section{Results}

\subsection{Distillation results}
After training or distillation, all agents were evaluated for $1 \times 10^{6}$ time steps of game play. Depending on the agent and game, $1 \times 10^{6}$ time steps resulted in 4 to 56 episodes per game. Results are given in Table\,\ref{tab:avg_scores}. The bottom two rows of the table provide the geometric mean of the student versus the geometric mean of the PPO-based teacher and the geometric mean of DQN-based teacher scores reported in \cite{rusu2015policy}. PPO is a more advanced algorithm than DQN, and even our low capacity PPO-trained agent obtain scores much higher than the DQN teacher.

The capacity of a student has an impact on how much information is transferred to a student. The medium capacity students obtain a geometric mean of 94\% relative to the teacher, and the low capacity students obtain 85\%. In general, then, it is beneficial to use larger capacity students.

\textbf{Critically, as given in the Medium vs. Medium AD and Low vs. Low AD columns in Table\,\ref{tab:avg_scores}, distilled students significantly exceed or meet the performance of equal-capacity agents which were trained directly on the environment.} Recall from Table\,\ref{fig:score-cap2} that higher capacity policy networks typically reach higher performance, faster than lower capacity networks. By using distillation we may exploit this fact and not be penalized by it.

\subsection{Effect of distillation epochs}
The optimal number of epochs used for distillation depends on the environment. Some games, e.g. Pong and Freeway, required 10 epochs of distillation to reach teacher performance. Others, e.g. Breakout and Ms.Pacman, required hundreds of epochs. The effects of increasing the number of distillation epochs on student evaluation performance for four games is given in Fig.\,\ref{fig:epochs}. In general, higher capacity policies distill with higher final evaluation performance than lower capacity policies, but the performance difference diminishes as the number of epochs increase. 

Each data point in Fig.\,\ref{fig:epochs} was created by initializing a new student policy (with random weights) and then distilling from between 10 and 400 epochs\footnote{Beamrider, Breakout, Enduro, and Riverraid were the only students distilled for 400 epochs.}, and then finally evaluating the distilled student for $1 \times 10^6$ time steps in the environment. For the sake of sample efficiency, it would be preferable to have access to a proxy metric to know when further distillation is unnecessary, but we leave that for future work. 

\subsection{Fine-tuning results}
We also studied the impact of allowing distilled students to learn in the environment, using standard PPO, after the distillation phase. Students were distilled for 10 epochs and then fine-tuned for $20 \times 10^6$ time steps, which is 27\% of the number of time steps used to directly train the medium and low capacity policies. \textbf{Notably, fine-tuning elevated the performance of the medium capacity distilled student to the performance of the teacher.} In Table\,\ref{tab:avg_scores}, the ``Medium AD tuned'' and ``Low AD tuned'' students have geometric means significantly higher than the students which were only distilled and not fine-tuned.  

\section{Conclusions}
Distillation is a robust and generally applicable optimization method. In this paper we show that distillation may be used successfully in conjunction with Proximal Policy Optimization, a popular actor-critic reinforcement learning algorithm. The method presented here can be used during architecture search for efficient hardware and policy co-design. 

Specifically, a high capacity trained teacher may be used to collect a replay buffer of state observations from the environment. Then the replay buffer and teacher probabilities may be used repeatedly to experiment with different student architectures. This method trains a low capacity reinforcement learning policy to achieve higher performance than it would have through direct interaction with the environment. 

Furthermore, if it is possible for the student to also learn within the environment, we show that it is beneficial to first perform distillation followed by fine-tuning of the student.

Distillation of policies was originally in the context of a neural network trained to approximate Q-values. A limitation of Q-values is the inability to represent action values for continuous action spaces. Actor-critic methods have no such limitation. Future work can extend the ideas here to continuous action spaces.

The field of deep learning has a training heuristic called \textit{early stopping}, which can be used to prevent overfitting. Early stopping monitors error on the training set, relative to error on a test dataset. As epochs increase, training error will always decrease, however test error will reach a minimum, before increasing again. Early stopping may be beneficial for distillation, but it is not clear. As may be seen in Fig.\,\ref{fig:epochs}, distilled student performance is plateauing, but generally not dropping as epochs increase. We leave further investigation into this question for future work.

\section*{Acknowledgment}
This work was supported by the DOE Advanced Simulation and Computing program, and the Laboratory Directed Research and Development program at Sandia National Laboratories. Sandia National Laboratories is a multi-program laboratory managed and operated by National Technology and Engineering Solutions of Sandia, LLC., a wholly owned subsidiary of Honeywell International, Inc., for the U.S. Department of Energy's National Nuclear Security Administration under contract DE-NA-0003525. This paper describes objective technical results and analysis. Any subjective views or opinions that might be expressed in the paper do not necessarily represent the views of the U.S. Department of Energy or the United States Government.

\bibliographystyle{abbrv}
\bibliography{PPO-IEEEtran}

\end{document}